%% file: main.tex
\documentclass{article} 
\usepackage{iclr2027_conference,times}

\input{math_commands.tex}

\usepackage[T1]{fontenc}
\usepackage[utf8]{inputenc}
\usepackage{latexsym}
\usepackage{microtype}
\usepackage{inconsolata}
\usepackage{graphicx}
\usepackage{booktabs}
\usepackage{multirow}
\usepackage{pifont}
\usepackage{xcolor}
\usepackage{comment}
\usepackage{xspace}
\usepackage{pgfplots}
\usepgfplotslibrary{groupplots}
\pgfplotsset{compat=1.18}
\usepackage{hyperref}
\usepackage{url}
\definecolor{darkblue}{rgb}{0, 0, 0.5}
\hypersetup{colorlinks=true, citecolor=darkblue, linkcolor=darkblue, urlcolor=darkblue}

\newcommand{\koffseq}{KOFF:~sequential\xspace}
\newcommand{\koffjoint}{KOFF:~joint\xspace}

\title{Knowledge Offloading:  Decomposing LLMs into Sparse Backbones and Memory Modules}

\author{
Karim Galliamov\textsuperscript{1}
\quad
Rochelle Choenni\textsuperscript{1} 
\quad
Ivan Titov\textsuperscript{2,1} \\
\textsuperscript{1}University of Amsterdam
\quad
\textsuperscript{2}University of Edinburgh \\
\texttt{karim.galliamov@student.uva.nl}
\quad
\texttt{\{r.m.v.k.choenni, titov\}@uva.nl}
 }

\iclrfinalcopy 

\begin{document}

\maketitle
\lhead{Preprint}
\vspace{-12pt} 

\begin{abstract}
LLMs encode both general capabilities and domain-specific knowledge in a single set of parameters. We ask whether this capacity can be reorganized: keeping broadly useful computation in a shared backbone, while moving specialized knowledge into external memory modules. We propose \emph{knowledge offloading} (KOFF), a framework for decomposing a pretrained LLM into a sparse shared backbone and domain-specific memories. Starting from a frozen base model, we either jointly or sequentially learn a structured pruning mask and lightweight recovery modules, implemented as LoRA adapters and learned key-value caches.  Across Llama and Qwen models from 3B to 14B, we find that 12\% to 20\% of capacity can be moved out of the shared backbone while largely preserving perplexity and MMLU performance (up to 90\% of the full model).
We show that LoRA and learned KV memories are complementary, and analyses suggest that the offloaded knowledge is specialized: mismatched memories do not yield the same gains, and language-specific neurons are preferentially removed while language-neutral neurons largely remain in the backbone. These results suggest that the knowledge of a pretrained LLM can be reallocated between a shared core and swappable external memories.\footnote{Code for reproducing the experiments will be available \href{https://anonymous.4open.science/r/MemoryOffloading-847F/README.md}{here}.}
\end{abstract}

\section{Introduction}

Large language models (LLMs) are expected to perform in diverse settings: a single model may need to serve users in multiple languages and reason across different domains. These use cases place high demands on model capacity. Some knowledge is broadly useful, supporting general reasoning and linguistic competence, while other knowledge is highly specialized, relevant only to particular topics or languages. Yet, LLMs lack explicit mechanisms to manage how capacity is allocated across different types of knowledge. All information, both general and highly specialized, is stored within a single set of parameters, making it difficult to update, remove, or extend specialized knowledge without risking interference with unrelated capabilities. 
This motivates a modular perspective~\citep{pfeiffermodular}, in which knowledge is organized into separate, manageable components rather than a single undifferentiated set of parameters.
Thus, we raise a fundamental question: can we reorganize model capacity so that broadly useful knowledge remains in a shared core, while specialized knowledge is moved into external modules that can be managed independently of the core?

A natural approach is to augment LLMs with external memory. Prior work has explored retrieval-augmented generation~\citep{lewis2020rag}, learned KV caches~\citep{eyuboglu2025cartridges}, and parameter-efficient adapters~\citep{hu2022lora} as mechanisms to extend model capacity. These approaches allow specialized knowledge to be stored outside the main parameters and accessed when needed, reducing the burden on the shared model. However, such methods typically assume a fixed backbone and do not decide what should remain in the model and what should be externalized. As a result, they address how to add memory, but not how to relocate capacity between shared computation and specialized knowledge. In contrast, structured pruning methods provide a way to reshape model capacity by removing parameters deemed less important~\citep{ma2023llmpruner}. Yet, pruning methods are designed to produce a single compressed network that serves all inputs, forcing specialized knowledge from different domains to compete for limited capacity.

\begin{figure}[t!]
    \centering
    \scalebox{0.6}{
    \includegraphics[width=\linewidth]{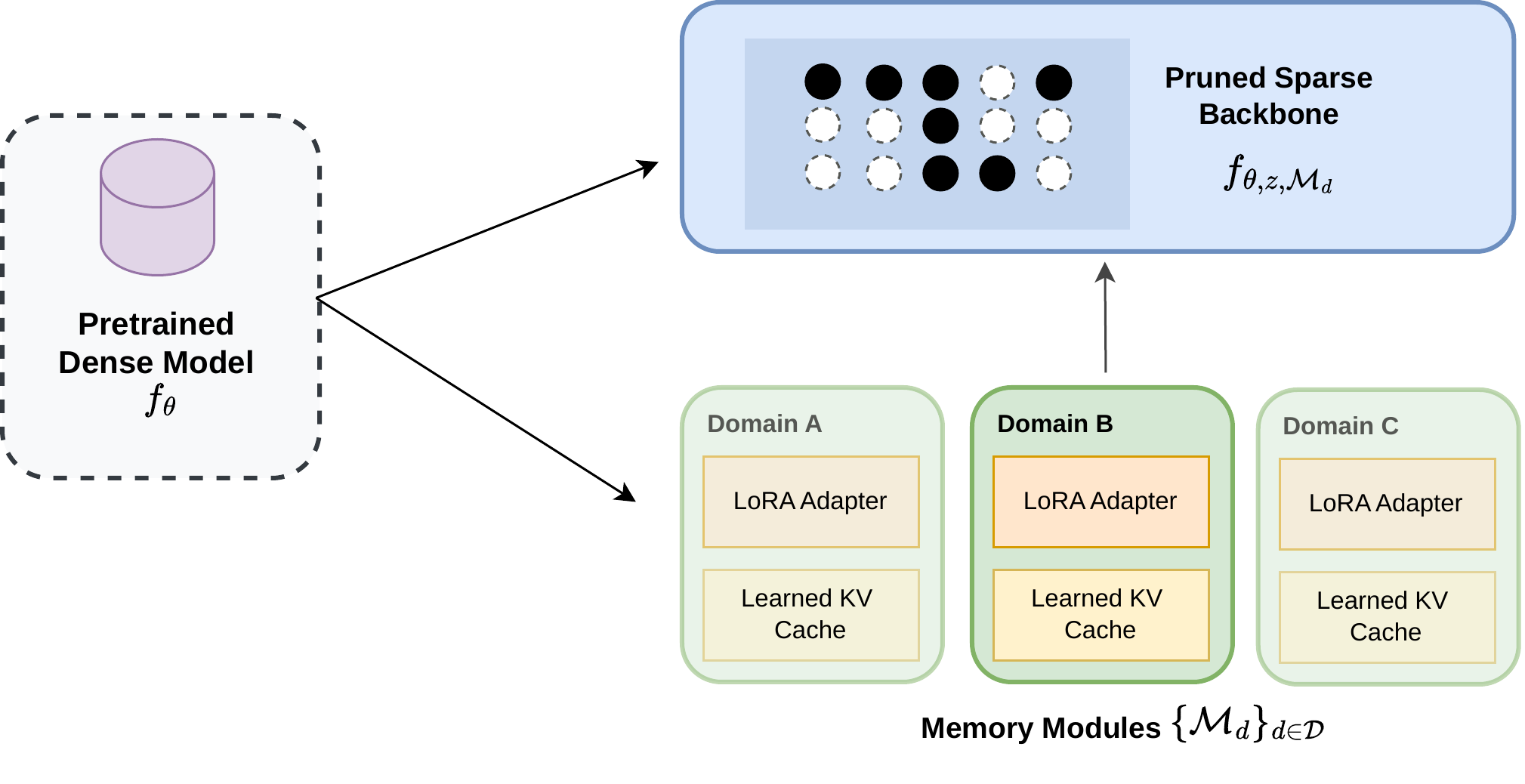}}
    \vspace{-0.3cm}
    \caption{Knowledge offloading  (KOFF) decomposes a pretrained LLM into a sparse backbone and separate swappable memory modules.}
    \vspace{-0.5cm}
    \label{fig:main_figure}
\end{figure}

In this work, we propose \emph{Knowledge Offloading} (KOFF), a framework that unifies memory-augmented modeling and structured pruning by learning how to relocate capacity between shared computation and external memory. Instead of compressing a model into a single smaller network, KOFF decomposes a pretrained LLM into a shared sparse backbone and domain-specific memory modules: a structured pruning mechanism based on Hard Concrete gates~\citep{louizos2018learning} determines which neurons are retained in the backbone, while memory modules are trained to compensate for the removed capacity through two recovery mechanisms: (1) a LoRA adapter~\citep{hu2022lora} operating in parameter space, and (2) a learnable key-value (KV) cache injected into the attention mechanism~\citep{eyuboglu2025cartridges}, acting as an externalized knowledge store. We learn this decomposition either sequentially (\koffseq), pruning first and then training the modules on the frozen backbone, or jointly (\koffjoint), so that the mask can account for which capacity the modules can recover. Comparing the two lets us test whether deciding what to offload benefits from knowing what the modules can recover. 
At inference time, a small trained classifier selects the memory module to attach to the backbone for a given input.

We test KOFF on 4 LLMs ranging in size from 3B to 14B, using topic and language domains. Note that our goal is not to develop a more efficient pruning method instead, we use pruning without memory modules as a control. If specialized knowledge can be offloaded, attaching the matched memory module to a sparse backbone should recover more of the full model's performance than the backbone alone. Our results support this, at 20\% sparsity, attaching the matched memory module consistently lowers perplexity relative to the pruning-only baseline and often improves MMLU, even when domains of different types are mixed. On Global MMLU, pruning alone already retains most of the full model's performance while perplexity rises substantially, consistent with the backbone retaining knowledge shared across languages. Neither variant consistently dominates: \koffjoint is better on Llama-3.1-8B, while \koffseq gives lower perplexity on Qwen models. For the instruct model, language modules trained on raw text reduce perplexity but hurt Global MMLU, showing that the modules offload whatever the domain data represents, including its format. Finally, ablations show that LoRA adapters and learned KV caches are complementary recovery mechanisms.

Importantly, recovering performance is a necessity but does not show that recovered knowledge is actually domain-specific. Our specialization analyses indicate that the decomposition reflects the intended split between shared and specialized knowledge. Functionally, the modules are domain-specific: attaching a mismatched module yields worse perplexity than the matched one. Structurally, the pruning mask removes most language-specific neurons from the backbone, while language-neutral neurons are almost entirely retained. Together, these results suggest that the capacity of a pretrained LLM can be reorganized into a shared core and separately managed specialized modules.  
To our knowledge, this is the first work to treat pruning as a mechanism for learning memory allocation.

\section{Related Work}

\paragraph{Memory-augmented models.}

Prior work extends LLM capacity by externalizing knowledge in activation or parameter space: prefix and prompt tuning inject learnable representations into the model's computation~\citep{li2021prefix,lester2021prompt}, while LoRA~\citep{hu2022lora} adds lightweight parameter updates. Drawing inspiration from these works, we let the model learn how to offload knowledge to such external memory. LoRA adapters provide parameter-space memory, while KV caches act as activation-space memory. Moreover, in multilingual NLP, modularity has been achieved by adding language-specific adapters to a frozen model~\citep{pfeiffer2020mad}, or by identifying language subnetworks through pruning~\citep{foroutan2022discovering, choenni2024examining}. KOFF combines both, removing language-specific capacity from the backbone and relocating it to external modules. 
Recent work also externalizes factual knowledge into editable databases during pretraining~\citep{zhao2025pre}, trains compressed KV caches offline via context distillation~\citep{eyuboglu2025cartridges}, and shows that sparse memory-layer updates enable continual learning without catastrophic~\citep{lin2025continual}. Closest to our work are parametric memory approaches such as memory networks~\citep{sukhbaatar2015endtoend} and hierarchical parametric memory~\citep{pouransari2025hierarchical}, which learn memory alongside the model or from scratch.  We instead start from a pretrained dense model, prune it, and use memory modules to recover knowledge lost during pruning. Unlike prior work, we treat memory as the recovery mechanism, where pruning and memory can be learned in sequence or jointly.

\paragraph{Model pruning.}
Prior work has shown that LLMs can be compressed via post-training structured pruning. 
Popular methods such as SliceGPT~\citep{ashkboos2024slicegpt}, FASP~\citep{hu2025fasp}, and LLM-Pruner~\citep{ma2023llmpruner} remove entire neurons or channels while retaining most of the base model's performance. The latter popularized the prune-then-adapt paradigm, where a pruned model is subsequently recovered using parameter-efficient methods such as LoRA. However, these approaches produce a single model with a fixed quality--efficiency trade-off.
We instead use structured pruning to reorganize model capacity: by learning a shared sparse backbone together with domain-specific memory modules, we allow different pruned knowledge to be restored for different inputs. This reframes pruning from a purely compressive operation into a component of a broader memory allocation strategy. Our sequential variant can be read as a domain-wise generalization of prune-then-adapt, recovering the removed capacity into several swappable modules rather than into one model, while our joint variant additionally lets the pruning mask depend on what those modules can recover.

\paragraph{Conditional computation and expert models.}
Dense-to-MoE conversion and expert-LM methods: MoEfication \citep{zhang-etal-2022-moefication}, upcycling~\citep{komatsuzaki2023sparse}, Branch-Train-Merge~\citep{li2022branchtrainmerge} and Branch-Train-MiX~\citep{sukhbaatar2024branchtrainmix} introduce conditional specialization by constructing or routing expert computation. Our pruning objective instead makes the allocation of capacity itself the object of learning, and our modules are memory components attached to a shared backbone rather than routed FFN experts.

\section{Methodology}
\label{methodology}

\subsection{General framework}
\label{sec:general-framework}

KOFF relies on a separation of roles: the shared backbone preserves computation that is broadly useful across domains, while domain modules provide capacity for domain-specialized information. We consider two ways of learning this separation. In \koffjoint, the pruning mask is trained jointly with the domain modules. This gives the pruning objective access to the same recovery mechanisms that will be available after pruning: capacity that can be restored by a memory module need not be retained in the shared backbone, while capacity that is difficult to recover, or useful across many domains, should remain shared. In \koffseq, the mask is first learned without modules and then frozen, after which the modules are trained on top of the fixed backbone. This is simpler and decouples the two stages, but the mask cannot anticipate what the modules will be able to recover. Each module \(\mathcal{M}_d\) contains two complementary forms of memory: a parameter-space component, implemented with LoRA, and an activation-space component, implemented as a learned key-value memory. These components are described in \S~\ref{sec:realization} and are both shown beneficial in \S~\ref{sec:component-ablation}.

\paragraph{Learning objective.}
Let \(f_\theta\) be the original pretrained model, with frozen parameters \(\theta\). We learn a shared structured pruning mask \(z\), together with domain modules \(\{\mathcal{M}_d\}_{d \in \mathcal{D}}\). The mask \(z\) is shared across domains, while each module \(\mathcal{M}_d\) is domain-specific. We write \(f_{\theta,z}\) for the masked backbone, and \(f_{\theta,z,\mathcal{M}_d}\) for the model obtained by attaching the module for domain \(d\).

Our goal is to preserve the behavior of the original model while moving recoverable capacity into external memory. We therefore use distillation from \(f_\theta\), rather than standard next-token cross-entropy, on the domain data. At a high level, the objective is
\begin{equation}
\min_{z,\{\mathcal{M}_d\}}\;
\sum_{d\in\mathcal{D}}
\;\mathbb{E}_{x\sim \mathcal{D}_d}
\bigl[
\mathcal{L}_{\mathrm{distill}}(f_{\theta,z,\mathcal{M}_d}, f_\theta; x)
\bigr]
+\; \lambda\,\Omega(z).
\label{eq:total-loss}
\end{equation}
The first term trains each domain module to recover the teacher's behavior on its own domain data, while updating the same shared mask \(z\). The penalty \(\Omega(z)\) encourages the shared backbone to use fewer active units; in practice, we instantiate it as an expected \(L_0\) penalty on structured Hard Concrete gates (\S~\ref{sec:sparse-backbone}). \koffjoint optimizes Eq.~\ref{eq:total-loss} directly. \koffseq splits it into two stages: the mask is first learned with no modules attached, minimizing the same objective with \(f_{\theta,z}\) in place of \(f_{\theta,z,\mathcal{M}_d}\); then \(z\) is frozen and each module is trained with the distillation term alone. 

\paragraph{Inference.}
The masked backbone need not be served as a dynamically gated model. After training, \(z\) is fixed and materialized: pruned neurons and their associated weight rows and columns are removed, yielding a compact dense backbone. At inference time, the domain label is not assumed to be known. We therefore train a lightweight router \(r_\phi(d \mid x)\) on the same domain-labelled data used to train the memories. Given an input \(x\), the router selects
$\hat d = \arg\max_{d\in\mathcal{D}} r_\phi(d \mid x)$,
and generation proceeds with \(f_{\theta,z,\mathcal{M}_{\hat d}}\). The router only selects which memory to attach; it does not change the shared mask or the compact backbone. We describe it further in Appendix~\ref{app:inference_router}.

\subsection{Domain-specific memory modules}
\label{sec:realization}

Each module $\mathcal{M}_d$ adds domain-specific capacity to the shared sparse backbone through two complementary mechanisms: a LoRA adapter and a learnable KV cache trained jointly.

\paragraph{LoRA adapters.}
For each domain (or language) $d \in \mathcal{D}$, we introduce a LoRA adapter \citep{hu2022lora}. The adapter provides a lightweight parameter-space correction to the pruned backbone. In our experiments, LoRA adapters target selected projection matrices and may be placed either in all transformer blocks or only in a contiguous subset of layers, such as the top $N$ blocks. We compare these options in Appendix~\ref{app:placement}.

\paragraph{Learnable KV cache.}
Alongside the LoRA adapter, each domain $d$ is equipped with a set of $N_{kv}$ learnable key-value memory representations at each attention layer where the module is active. As in \citet{eyuboglu2025cartridges}, these keys and values are not produced by Transformer computation; they are trainable parameters directly inserted into the attention mechanism. Specifically, at layer $l$ the memory consists of
$K_{d}^{(l)}, V_{d}^{(l)} \in \mathbb{R}^{h \times N_{kv} \times d_h}$,
where $h$ is the number of KV heads and $d_h$ is the head dimension. The memory keys and values are prepended to the regular text keys and values:
\begin{equation}
  \tilde{K}^{(l)} = [K_{d}^{(l)};\; K_{\text{text}}^{(l)}], \;
  \tilde{V}^{(l)} = [V_{d}^{(l)};\; V_{\text{text}}^{(l)}].
  \label{eq:kv-concat}
\end{equation}
The causal mask is extended so that every text position can attend to all $N_{kv}$ memory tokens, while the standard causal constraint is preserved among text positions.

\subsection{Learning a sparse backbone}
\label{sec:sparse-backbone}

We learn the shared mask using the $L_0$ regularization framework of
\citet{louizos2018learning}. Each removable channel has a stochastic binary
mask variable $z_i \in \{0,1\}$, relaxed during training with the Hard Concrete
distribution. The mask parameters define the distribution over the choice of retained
channels, and are not input-dependent gates. This lets us optimize a structured
sparsity pattern jointly or sequentially with the domain modules, then fix and materialize it
after training.

For a linear map with output dimension $n_{\text{out}}$, we learn
$z\in\{0,1\}^{n_{\text{out}}}$ over output channels. Let
$\boldsymbol{\alpha}$ denote the Hard Concrete parameters. We penalize the
expected number of retained channels,
$
  \mathcal{R}(\boldsymbol{\alpha})
  =
  \mathbb{E}_{\mathbf{z}}\bigl[\|\mathbf{z}\|_0\bigr],
$
and impose a target sparsity $s^{*}$ through a hinge penalty on the active
fraction:
\[
  \Omega(z)
  =
  \bigl[
    \hat a - (1-s^{*})
  \bigr]_{+},
  \qquad
  \hat a =
  \frac{\mathcal{R}(\boldsymbol{\alpha})}{|\boldsymbol{\alpha}|}.
\]
Here $[\cdot]_+=\max(0,\cdot)$. The loss encourages the shared backbone to keep
at most the target fraction of active channels, while the domain modules learn
to recover capacity that can be externalized. LoRA adapters modify the same linear maps that are being pruned, thus, the
mask must define which output channels remain available to both the pretrained
weights and the domain-specific correction. For a base linear map $W$ and a
domain-specific LoRA update $\Delta W_d$, we therefore apply the mask after the
two branches are added:
\[
  y = z \odot (W x + \Delta W_d x).
\]
This prevents the LoRA branch from reintroducing channels that have been removed
from the shared backbone, and makes the final structure explicit: intuitively, after training,
channels with $z_i=0$ can be removed from both the base map and the corresponding
LoRA output dimensions.

\subsection{Optimization protocol}
\label{sec:optimization}

Section~\ref{sec:general-framework} defines the objective; we now describe the training procedure. The pretrained parameters $\theta$ are frozen during training, we only optimize the shared mask and the domain-module parameters. This isolates our question: whether existing capacity can be relocated from the backbone into memory modules while retaining the behaviour of the original model. Domain training uses a token-level distillation loss from the original unpruned model to the student, equivalently $\mathrm{KL}(p_\theta(\cdot|x)\,\|\,p_{\theta,z,\mathcal{M}_d}(\cdot|x))$.
For each input, we store the teacher's top-100 logits and train the offloaded model to match the corresponding distribution. In \koffjoint, for a domain batch $(x,d)$, we attach the corresponding module $\mathcal{M}_d$ and update the shared mask together with that module, including the sparsity penalty. In \koffseq, the first stage updates only the mask, with no module attached; the second stage freezes the mask and updates only $\mathcal{M}_d$. Both variants are given the same budget: one pass over the domain corpus in total. \koffjoint uses it for a single joint pass, while \koffseq splits the corpus evenly, so its pruning stage and its module-training stage each see half of it. Differences between the two thus reflect the training schedule and not the amount of data or compute.

\section{Experiments}
\label{sec:experiments}

\label{subsec:settings}

\paragraph{Models}
We use 3 pretrained base models: Qwen-2.5-3B, Qwen-2.5-14B~\citep{qwen2025qwen25}, and Llama-3.1-8B~\citep{grattafiori2024llama3herdmodels}. We additionally report Llama-3.1-8B-Instruct to test whether the decomposition carries over to an instruction-tuned model.

\paragraph{Domain partitioning}
We use two different types of domains. First, \textbf{topic domains}: we use Wikipedia-Topics~\citep{tarekziade_wikipedia_topics}, which partitions English Wikipedia articles into 6 topic domains: Mathematics, Science, Technology, History, Culture, and Geography.  We sample 30{,}000 articles per domain, split each article into chunks of at most 512 tokens, and apply a 95/5 train/evaluation split (approximately 200M tokens in total). As a second partitioning strategy we use \textbf{language domains}. Specifically, we use Wikisource~\citep{wikimedia} and select 6 languages with sufficient coverage: English, Spanish, French, German, Italian, and Portuguese.  Each language is treated as a separate domain with its own memory module; documents are chunked in the same way, and the total volume of the training data is again roughly 200M tokens.

\paragraph{Evaluation}
We use two metrics: (i)~per-domain held-out perplexity, and
(ii)~MMLU~\citep{hendrycks2021measuring} (5-shot) for broad factual knowledge; for language domains we use Global MMLU~\citep{singh2025globalmmluunderstandingaddressing} restricted to the six training languages, each evaluated with its matched module. Some experiments in the Appendix additionally report BigBench-Lite (BBL)~\citep{srivastava2023beyond}.

\paragraph{Baselines} 

We evaluate both KOFF variants: \koffjoint, which learns the mask and memory modules together, and \koffseq, which first learns the mask, freezes it, and trains the modules on top. Both are trained using the same budget and we compare them against two baselines. \emph{Pruning-only} learns a mask in the same way, but without memory modules. This allows us to test how much of the removed capacity the KOFF modules recover. Secondly, we use LLM-Pruner~\citep{ma2023llmpruner} with a recovery stage, an established structured pruning method, as an external reference point. For recovery, we use the same slice of Wikipedia used to train the other models.

\paragraph{Pruning and module configuration}  We prune the attention projections ($W_K$, $W_V$) 
together with the MLP projections ($W_{\text{gate}}$, $W_{\text{up}}$, $W_{\text{down}}$), this is also where LoRA adapters are applied. 
We use LoRA rank 128 and 512 KV memory tokens in each setting. Moreover, we found that attaching  memory modules to higher layers works best, thus we always attach them to the last 16 transformer blocks. Finally, we train one memory module per domain, i.e. 6 in each setting, with each module adding roughly 1--3\% of the base model's parameters. At inference, when only one module is loaded, its LoRA adapter are merged into the pruned weights, so only the KV memory adds parameters. Thus, the backbone with one module is about 90\% of the original size at 12\% sparsity and 83\% at 20\%. For parameter counts and hyperparameters, see Appendix~\ref{app:implementation}.

\section{Results}
\label{sec:results}

We ask whether a pretrained LLM's capacity can be reorganized so that broadly useful knowledge stays in a shared backbone while specialized knowledge moves to separate modules. A necessary condition is that attaching the matched memory module to a sparse backbone recovers more of the full model's performance than the backbone alone. Thus, we test this on Llama-3.1-8B (\S\ref{sec:main-results}), across model sizes (\S\ref{sec:scale}), and on an instruction-tuned model (\S\ref{sec:instruct}), and we ablate the memory modules and test mixed domains (\S\ref{sec:component-ablation}). Whether the recovered knowledge is domain-specific is examined in \S\ref{sec:analysis}.

\begin{table*}[t!]
\centering
\footnotesize
\setlength{\tabcolsep}{4pt}
\scalebox{0.9}{
\begin{tabular}{l | cc | cc}
\toprule
 & \multicolumn{2}{c}{Topic domains}
& \multicolumn{2}{c}{Language domains} \\
\cmidrule(r){2-3}\cmidrule(l){4-5}
Model 
& MMLU$\uparrow$ & PPL$\downarrow$
& Global MMLU$\uparrow$ & PPL$\downarrow$ \\
\midrule
Full Llama-8B (unpruned)     & 0.667 &  7.93 & 0.541 &  9.70 \\
\midrule
LLM-Pruner        & 0.471 &  9.98 & 0.396 & 18.72 \\
\midrule
Pruning only     & 0.400 & 10.88 & 0.532 & 14.03 \\
\koffseq        & 0.592 &  9.13 & 0.534 & 11.95 \\
\koffjoint      & \textbf{0.602} &  \textbf{8.12} & \textbf{0.537} & \textbf{10.90} \\
\bottomrule
\end{tabular}}
\caption{Topic- and language-domain results at 20\% global sparsity, Llama-8B base model. Pruning only uses the same $L_0$ sparsification and distillation objective but no memory modules. \koffseq learns a mask in a first stage and then trains the memory modules on top of it; \koffjoint learns the mask and the modules together. All models are trained for one pass over the domain corpus.
}
\vspace{-0.3cm}
\label{tab:main-results}
\end{table*}

\subsection{Main results}
\label{sec:main-results}

Our main setting is Llama-3.1-8B at 20\% global sparsity.\footnote{Unlike unstructured pruning, typically reported at 50\% sparsity, structured pruning is considerably harder and is studied at lower sparsity levels.} We use two domain partitions (\S\ref{sec:experiments}): topics (subject areas) and languages. The latter introduces a stronger distribution shift, as the modules must account for different token distributions as well as different content. We evaluate on MMLU for topics and Global MMLU for languages, and in both settings report perplexity on held-out domain text with the matched module attached, averaged across domains.

Table~\ref{tab:main-results} shows the results. In the topic setting, pruning alone substantially degrades perfromance: MMLU drops from 0.67 to 0.40 and perplexity rises from 7.9 to 10.9. Both KOFF variants recover most of this loss. \koffjoint reaches 0.60 MMLU and 8.12 perplexity. \koffseq reaches 0.59 and 9.1 on MMLU and perplexity respectively. In the language setting, perplexity shows the same trend: pruning-only results in 14.0 against 9.7 for the unpruned model, while \koffjoint and \koffseq reduce this to 10.9 and 12.0 respectively. Global MMLU, on the other hand, seems largely insensitive to this pruning setting. Pruning-only already retains 98\% of the full model's performance, and both KOFF variants close this gap even further. This result is consistent with the intended decomposition. The capacity removed from the backbone mainly affects language-specific modeling of the text, captured by perplexity, while the knowledge probed by Global MMLU, which is largely shared across languages, remains in the backbone. LLM-Pruner, the external baseline, scores worse than both KOFF variants in all four settings. 
Finally, on this model \koffjoint outperforms \koffseq across all settings, suggesting that deciding what to remove from the backbone benefits from knowing what the modules can recover.

\begin{figure*}[t!]
\centering
\scalebox{0.72}{
\begin{tikzpicture}
\begin{groupplot}[
    group style={group size=2 by 2, horizontal sep=1.5cm, vertical sep=1.8cm},
    width=0.42\textwidth, height=4.2cm,
    ybar=0pt, area legend,
    xtick={1,2,3}, enlarge x limits=0.25,
    xticklabels={Qwen-3B\\(12\%), Llama-8B\\(20\%), Qwen-14B\\(20\%)},
    xticklabel style={align=center, font=\scriptsize},
    yticklabel style={font=\scriptsize},
    title style={font=\small, yshift=-4pt},
    ylabel style={font=\scriptsize},
    ymajorgrids, grid style={gray!20},
    tick align=outside, axis x line*=bottom, axis y line*=left,
]
\nextgroupplot[title={Topic: MMLU}, ylabel={\% retained}, ymin=40, ymax=102,
    extra y ticks={100}, extra y tick labels={},
    extra y tick style={grid=major, major grid style={dashed, black!60}},
    legend to name=scalelegend, legend columns=4,
    legend style={font=\scriptsize, draw=none, /tikz/every even column/.append style={column sep=6pt}}]
\addplot[fill=gray!50, draw=none]  coordinates {(1,44.5) (2,60.0) (3,94.2)};
\addplot[fill=blue!40, draw=none]  coordinates {(1,81.4) (2,88.8) (3,95.6)};
\addplot[fill=blue!80!black, draw=none] coordinates {(1,72.8) (2,90.3) (3,95.9)};
\addplot[fill=orange!70, draw=none] coordinates {(1,92.6) (2,70.6) (3,90.4)};
\legend{Pruning only, \koffseq, \koffjoint, LLM-Pruner}
\nextgroupplot[title={Topic: PPL}, ylabel={\% increase}, ymin=0, ymax=40]
\addplot[fill=gray!50, draw=none]  coordinates {(1,10.6) (2,37.2) (3,21.1)};
\addplot[fill=blue!40, draw=none]  coordinates {(1,0.5) (2,15.1) (3,10.2)};
\addplot[fill=blue!80!black, draw=none] coordinates {(1,0.7) (2,2.4) (3,11.5)};
\addplot[fill=orange!70, draw=none] coordinates {(1,15.0) (2,25.9) (3,34.4)};
\nextgroupplot[title={Language: Global MMLU}, ylabel={\% retained}, ymin=40, ymax=102,
    extra y ticks={100}, extra y tick labels={},
    extra y tick style={grid=major, major grid style={dashed, black!60}}]
\addplot[fill=gray!50, draw=none]  coordinates {(1,43.3) (2,98.3) (3,96.1)};
\addplot[fill=blue!40, draw=none]  coordinates {(1,75.5) (2,98.7) (3,98.4)};
\addplot[fill=blue!80!black, draw=none] coordinates {(1,54.8) (2,99.3) (3,97.3)};
\addplot[fill=orange!70, draw=none] coordinates {(1,69.2) (2,73.2) (3,84.9)};
\nextgroupplot[title={Language: PPL}, ylabel={\% increase}, ymin=0, ymax=95]
\addplot[fill=gray!50, draw=none]  coordinates {(1,12.4) (2,44.6) (3,26.8)};
\addplot[fill=blue!40, draw=none]  coordinates {(1,2.6) (2,23.2) (3,6.4)};
\addplot[fill=blue!80!black, draw=none] coordinates {(1,4.6) (2,12.4) (3,7.4)};
\addplot[fill=orange!70, draw=none] coordinates {(1,42.5) (2,93.0) (3,45.7)};
\end{groupplot}
\end{tikzpicture}}

\ref{scalelegend}
\vspace{-0.1cm}
\caption{Results across model scales, relative to each model's unpruned version. For MMLU we show the percentage of the unpruned model's accuracy that is retained. For PPL we report the percentage increase in perplexity over the unpruned model (lower is better).}
\label{fig:scale}
\vspace{-0.2cm}
\end{figure*}
\subsection{KOFF across model sizes}
\label{sec:scale}

We now test whether the results hold across model sizes. Specifically, we test Qwen-2.5-3B at 12\% and Qwen-2.5-14B and Llama-3.1-8B at 20\% sparsity. Note that these models differ in sparsity. This is because the acceptable sparsity ratio depends on model size. For the 3B model, we found that model quality gradually degrades up to around 12\% sparsity, but then decreases sharply, with memory modules no longer compensating at 20\%. The 8B and 14B models were more resilient to larger sparsity ratios up to 20\%. Thus, to test non-trivial capacity relocation, we report the 3B model at 12\% and the larger models at 20\%. However, this makes absolute scores not directly comparable across models. Thus, in Figure~\ref{fig:scale} we report model performance relative to its own unpruned model version: the fraction of MMLU performance that is retained and the relative increase in perplexity.\footnote{For results in absolute numbers, see Appendix~\ref{app:full-results}.}

We find that KOFF improves over pruning-only perplexity for every model in both settings. Yet the extent of improvement seems largely dependent on how damaging pruning alone is. On Qwen-2.5-3B, pruning-only collapses MMLU to near random performance, even though the increase in perplexity stays modest. For this model, \koffseq recovers 81\% and 75\% of unpruned MMLU performance. This is also the only setting where LLM-Pruner outperforms KOFF (on MMLU only), although at the cost of higher perplexity. On Qwen-2.5-14B, by contrast, pruning alone already retains 94--96\% of MMLU and Global MMLU performance, leaving less room for recovery. Here KOFF gains are most prominent on perplexity evaluation. Finally, the preference between the two KOFF variants is model-dependent. \koffjoint is better on Llama-3.1-8B. \koffseq is better on Qwen-2.5-3B. On Qwen-2.5-14B the results are mixed, on MMLU  \koffjoint is slightly better and vice versa on Global MMLU, though differences are negligible. Across scales, the modules thus contribute most where pruning removes the most, when the backbone already retains most of the general capabilities, as in Qwen-2.5-14B, the modules mainly recover domain-specific modeling, reflected in perplexity rather than MMLU.

\subsection{Instruction-tuned models}
\label{sec:instruct}
In Table~\ref{tab:instruct-results}, we report KOFF results on Llama-3.1-8B-Instruct at 20\% sparsity. In the topic setting, we find that the results mirror those of the base model. KOFF improves over pruning-only, both in terms of MMLU and perplexity, with \koffjoint performing best in both cases. For languages, both KOFF variants again substantially reduce perplexity, however, unlike the base model, they underperform on Global MMLU compared to pruning-only. Given that the language modules are trained on raw monolingual text, we hypothesize that they shift the instruction-tuned model toward plain continuation behavior, which helps perplexity but interferes with instruction-following capabilities in non-English languages. This also supports the hypothesis that the modules offload whatever the domain data represents, including its format. Thus, for instruction-tuned models, the memory modules might need to be trained on data that preserves the instruction format, which we leave to future work.

\begin{table}[t!]
\centering
\small
\setlength{\tabcolsep}{4pt}
\scalebox{0.9}{
\begin{tabular}{l | cc | cc}
\toprule
& \multicolumn{2}{c}{Topic domains} & \multicolumn{2}{c}{Language domains} \\
\cmidrule(r){2-3}\cmidrule(l){4-5}
Method & MMLU$\uparrow$ & PPL$\downarrow$ & Global MMLU$\uparrow$ & PPL$\downarrow$ \\
\midrule
Unpruned           & 0.672 & 7.53 & 0.575 & 10.79 \\
\midrule
Pruning only       & 0.603 & 9.95 & \textbf{0.531} & 16.75 \\
\koffseq           & 0.618 & 9.37 & 0.445 & \textbf{11.70} \\
\koffjoint         & \textbf{0.636} & \textbf{9.07} & 0.459 & 12.21 \\
\midrule
LLM-Pruner         & 0.508 & 10.63 & 0.428 & 20.86 \\
\bottomrule
\end{tabular}}
\caption{Llama-3.1-8B-Instruct at 20\% global sparsity.}
\vspace{-0.1cm}
\label{tab:instruct-results}
\end{table}

\begin{table*}[t!]
\begin{minipage}[t]{0.36\linewidth}
\centering
\small
\begin{tabular}{lcc}
\toprule
& MMLU$\uparrow$ & PPL$\downarrow$ \\
\midrule
Unpruned                  & 0.67 & 7.9 \\
\midrule
Pruning only            & 0.40 & 10.9 \\
Pruning + LoRA          & 0.56 & 8.9 \\
Pruning + KV cache      & 0.58 & 8.5 \\
\koffjoint    & \textbf{0.60} & \textbf{8.1} \\
\bottomrule
\end{tabular}
\caption{Memory module ablation on Llama-3.1-8B at 20\% sparsity for topic domains. LoRA and KV cache are trained jointly during pruning.}
\label{tab:ablation-components}
\end{minipage}%
\hfill
\begin{minipage}[t]{0.52\linewidth}
\centering
\small
\begin{tabular}{lcccc}
\toprule
& \multicolumn{2}{c}{11 domains} & \multicolumn{2}{c}{3 domains} \\
\cmidrule(lr){2-3}\cmidrule(lr){4-5}
 & MMLU$\uparrow$  & PPL$\downarrow$ & MMLU$\uparrow$ & PPL$\downarrow$ \\
\midrule
Unpruned       & 0.667    & 7.94           & 0.541    & 7.22 \\
\midrule
Pruning only   & 0.607 & 11.78          & 0.485 & 11.12 \\
KOFF:seq.      & \textbf{0.638} & 10.68 & \textbf{0.511} & 10.10 \\
\koffjoint     & 0.630 & \textbf{10.21} & 0.504 & \textbf{9.71} \\
\bottomrule
\end{tabular}
\caption{KOFF on Llama-3.1-8B at 20\% sparsity using two mixed domain partitions: 11 domains contains data on all topics and languages (excl. English) and 3 domains includes Math, French and German.
}
\label{tab:matched-compute}
\end{minipage}
\vspace{-0.2cm}
\end{table*}

\subsection{Ablations}
\label{sec:component-ablation}

\paragraph{Memory modules}
Each KOFF memory module combines two components: a LoRA adapter on the backbone weights and a learned KV cache. We first ablate them to test whether both are necessary, training each component on its own jointly with the mask and comparing against the full module. We run this ablation on Llama-3.1-8B in the topic setting at 20\% sparsity. 
In Table~\ref{tab:ablation-components}, we see that each component alone already recovers most of the loss from pruning. Pruning only drops MMLU from 0.67 to 0.40 and raises perplexity from 7.9 to 10.9. The KV cache alone brings these to 0.58 and 8.5, and perhaps surprisingly, is slightly better than LoRA alone on both metrics (0.56 and 8.9 respectively). However, combining the two gives the best result, retaining 90\% of the original MMLU performance and increasing perplexity by only 2\%.

\paragraph{Domain partitions}
In our main experiments, each partition contains 6 domains of a single category, either topics or languages. In practice, we may need to cover different types of domains at once, and the number of domains determines how much capacity each module must hold and how much the backbone must share across them. We therefore test whether the decomposition, and the preference for \koffjoint on Llama-3.1-8B, holds under partitions that mix domain types and vary in size. Table~\ref{tab:matched-compute} reports two such partitions on Llama-3.1-8B at $\sim$20\% sparsity: 11 domains (6 topics and 5 languages) and a minimal mixed partition of 3 domains (Mathematics, French, German). Both KOFF variants improve over pruning only in both partitions: in the 11 domain mixture, perplexity drops from 11.78 to 10.68 (\koffseq) and 10.21 (\koffjoint); in the 3-domain mixture, from 11.12 to 10.10 and 9.71. The relative improvement over pruning only is nearly identical in both partitions (9\% for \koffseq and 13\% for \koffjoint), suggesting that a shared backbone can be learned across heterogeneous domains, and that the decomposition does not rely on domains being of a single type or on a particular number of domains.

 \section{Specialization analysis}

\label{sec:analysis}

In Section~\ref{sec:results} we have demonstrated the effectiveness of KOFF in terms of recovering model performance. We now delve a bit deeper and test our hypothesis that the learned decomposition produces a separation of roles, i.e. the shared backbone retains general knowledge and the memory modules specialize.  We first test to what extent memory modules are functionally specialized to their respective domains and we then look for structural specialization at the neuron level.

\begin{figure*}[t!]
\centering
\scalebox{0.72}{
    \centering
    \includegraphics[width=0.8\linewidth]{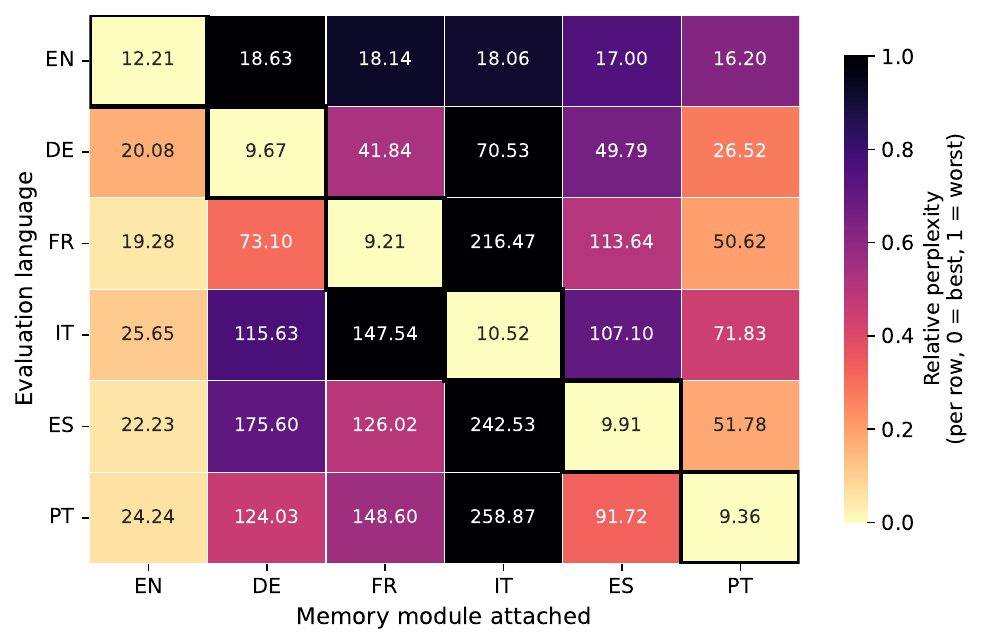}}
    \caption{Cross-language memory-module swap on Llama-3.1-8B (\koffjoint).
    We report perplexity scores when attaching the correct modules to the backbone (diagonal) versus mismatched ones.} 
    \label{fig:cross-domain-heatmap-lang}
\end{figure*}
\vspace{-0.1cm}

\paragraph{Modules specialize to domains}
\label{sec:cross-domain-swap}

We first test whether memory modules are functionally domain-specific. In the language-domain setting, we evaluate each language with every memory module attached to the shared backbone. If modules store domain-specific capacity, the matched module should give the best perplexity, while mismatched modules should degrade performance. Figure~\ref{fig:cross-domain-heatmap-lang} shows this pattern: perplexity is lowest on the diagonal, where the evaluation language and memory module match, and increases when the wrong module is attached. This suggests that the modules are not interchangeable adapters; they recover information specific to their own domains. Interestingly, the off-diagonal structure does not show a clear typological pattern: related languages do not consistently benefit from each other’s modules. For example, Romance-language modules are not uniformly better for other Romance languages, nor do English and German form a distinct cluster. For results on topic domains see Appendix~\ref{app:full-results}.

\paragraph{Language-specific capacity is preferentially offloaded}
\label{sec:lape-analysis}

Having shown that the modules specialize functionally, we ask whether the pruning mask shows a corresponding structural specialization. 
To do so, we use LAPE~\citep{tang2024language}, a method that scores neurons based on how language-specific their activation patterns are. We run LAPE on the training corpus of each language separately and score the MLP neurons of Llama-3.1-8B. We then classify the 15\% of neurons with the lowest entropy scores as language-specific and the 15\% with the highest entropy as language-neutral.
On Llama-3.1-8B model at 20\% sparsity, we find that roughly 70\% of the neurons that we classified as language-specific are removed from the backbone. In contrast, the language-neutral neurons are strongly protected: only $2.8\%$ are removed, even though the relevant projections are pruned at roughly $40\%$ local sparsity.
Thus, \koffjoint preferentially removes language-specific capacity while preserving language-neutral information. 
This is in line with our hypothesis that  memory modules recover offloaded language-specific information.

\section{Conclusion and Discussion}

We introduced knowledge offloading, which decomposes a pretrained LLM into a shared sparse backbone and swappable memory modules, implemented as LoRA adapters and learned KV caches, and learned either sequentially or jointly. Across models and domains, attaching the matched module preserves much of the unpruned model's MMLU performance and substantially reduces the perplexity increase caused by pruning, compared to a sparse backbone without modules. This suggests that the capacity of a pretrained LLM can be reorganized into a shared core and separately managed specialized modules. 
Our specialization analyses further support this idea: matched modules work best, and language-specific neurons are preferentially removed while language-neutral neurons are protected.

Yet, what is preserved depends on what the modules are trained for. Our objective distills token-level distributions on raw domain text, which is reflected in perplexity and MMLU but not sufficient for generative behavior. In preliminary experiments, short-form generation was less well preserved for all pruned variants, including those with memory modules, and language modules hurt Global MMLU on the instruction-tuned model. Preserving richer behavior such as generation, reasoning traces, tool use, and safety-relevant responses likely requires training objectives and data that explicitly target these behaviors, which we leave for future work.

\bibliographystyle{iclr2027_conference}
\bibliography{iclr2027_conference}
\appendix

\section{Implementation details}
\label{app:implementation}
Table~\ref{tab:hyperparams} lists the hyperparameters used in all main-text experiments. LoRA weights, gate parameters and KV memories use separate AdamW~\citep{loshchilov2018decoupled} optimizers. Models are loaded in bfloat16, and the 8B models use gradient checkpointing to fit in 80\,GB of GPU memory. The 8B and 14B models use a lower $\lambda$, to account for their larger number of gates, and a batch size of 1.

Table~\ref{tab:param-counts} gives parameter counts. With one module attached, the model is 10--18\% smaller than the original; the backbone together with all six modules is 88--107\% of the original size.

The Llama-3.2-3B design experiments in Appendices~\ref{app:distillation} and~\ref{app:placement} additionally use a small auxiliary term that periodically distills the teacher on general-domain text (C4~\citep{raffel2020exploring}, GSM8K~\citep{cobbe2021gsm8k} and ARC-Challenge~\citep{clark2018think}), updating only the mask. This term is not used in any main-text experiment.

\begin{table}[h]
\centering
\footnotesize
\begin{tabular}{ll}
\toprule
Parameter & Value \\
\midrule
LoRA rank $r$ / scaling $\alpha$   & 128 / 64 \\
LoRA target modules    & K, V, Gate, Down, Up \\
LoRA learning rate $\eta_{\text{LoRA}}$ & $3 \times 10^{-4}$ \\
Gate learning rate $\eta_{\text{gate}}$ & $2 \times 10^{-2}$ \\
Memory KV cache learning rate $\eta_{\text{mem}}$  & $1 \times 10^{-2}$ \\
L0 $\lambda$ (with auto-scaling) & 25 (60 for 3B) \\
Temperature $\beta$ & annealed from 2.0 to 0.5 \\
Memory tokens $N_{kv}$ & 512 per domain \\
Memory-module layers & last 16 blocks \\
Batch size   & 4 (1 for 8B or larger) \\
Max sequence length    & 512 \\
Epochs & 1 \\
\bottomrule
\end{tabular}
\caption{Default hyperparameters used in main-text experiments}
\label{tab:hyperparams}
\end{table}

\begin{table}[h]
\centering
\footnotesize
\begin{tabular}{lrrrrrrr}
\toprule
& & \multicolumn{2}{c}{Memory modules} & \multicolumn{2}{c}{Merged model} & \multicolumn{2}{c}{Backbone + all modules} \\
\cmidrule(lr){3-4}\cmidrule(lr){5-6}\cmidrule(lr){7-8}
Model & Base model & All domains & One domain & 12\% & 20\% & 12\% & 20\% \\
\midrule
Qwen2.5-3B   &  3.09B & 563.1M &  93.8M &  2.76B &  2.54B &  3.32B &  3.10B \\
Llama-3.1-8B &  8.03B & 906.0M & 151.0M &  7.21B &  6.65B &  8.10B &  7.54B \\
Qwen2.5-14B  & 14.77B & 950.0M & 158.3M & 13.20B & 12.14B & 14.13B & 13.07B \\
\bottomrule
\end{tabular}
\caption{Parameter counts of the base models, the KOFF memory modules, and the merged model used at inference. The merged model is the pruned backbone with one domain's module attached: the LoRA adapter is merged into the remaining weights and adds no inference-time parameters, while the KV cache is kept. Llama-3.1-8B-Instruct has the same architecture as Llama-3.1-8B, so its counts are identical.}
\label{tab:param-counts}
\end{table}

\begin{figure*}[t!]
    \centering
    \includegraphics[width=0.7\linewidth]{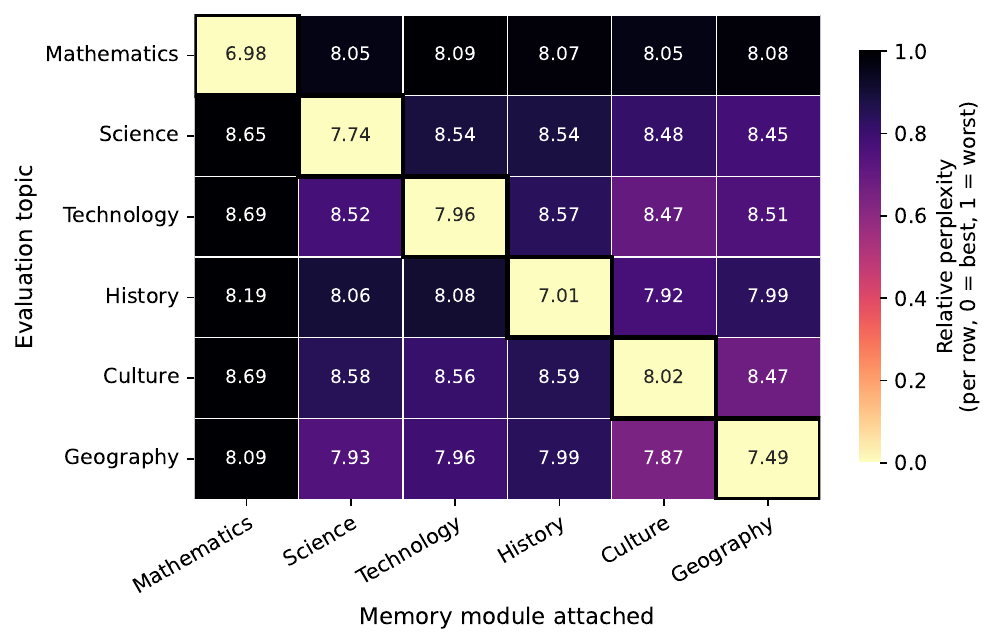}
    \caption{Cross-topic memory-module swap on Llama-3.1-8B ($\sim$12\% global sparsity).
    Each cell reports held-out perplexity on the row's evaluation topic when the column's memory
    module is attached to the shared backbone; the boxed diagonal is the matched module}
    \label{fig:cross-domain-heatmap-topic}
\end{figure*}

\subsection{Domain classification model for inference routing}
\label{app:inference_router}
We train a RoBERTa~\citep{liu2019roberta} on the same dataset as the main model (Wikipedia-Topics for topic-domain runs, WikiSource for language-domain runs) to classify texts into domains. All parameters are unfrozen, the learning rate is set to $0.0003$, and the batch size to 64. On a held-out test set the router's accuracy is 0.9675.

\section{Full results}
\label{app:full-results}

Table~\ref{tab:main-and-lang-results} gives the absolute scores behind Figure~\ref{fig:scale} and Table~\ref{tab:instruct-results}, together with Llama-3.1-8B at 12\% sparsity. 
Figure~\ref{fig:cross-domain-heatmap-topic} is the topic-domain counterpart of the cross-language module swap in Figure~\ref{fig:cross-domain-heatmap-lang}.

\begin{table*}[t!]
\centering
\setlength{\tabcolsep}{4pt}
\scalebox{0.9}{
\begin{tabular}{l c | cc | cc}
\toprule
& & \multicolumn{2}{c}{Topic domains}
& \multicolumn{2}{c}{Language domains} \\
\cmidrule(r){3-4}\cmidrule(l){5-6}
Model & Sparsity
& MMLU$\uparrow$ & PPL$\downarrow$
& Global MMLU$\uparrow$ & PPL$\downarrow$ \\
\midrule
Qwen-3B (unpruned)          & 0\%  & 0.633 & 8.21 & 0.571 & 13.12 \\
Qwen-3B (LLM-Pruner)        & 12\% & 0.586 & 9.44 & 0.395 & 18.69 \\
Qwen-3B (pruning only)      & 12\% & 0.282 & 9.08 & 0.247 & 14.75 \\
Qwen-3B (\koffseq)          & 12\% & 0.515 & 8.25 & 0.431 & 13.46 \\
Qwen-3B (\koffjoint)        & 12\% & 0.461 & 8.27 & 0.313 & 13.72 \\
\midrule
Llama-8B (unpruned)         & 0\%  & 0.667 &  7.93 & 0.541 &  9.70 \\
Llama-8B (LLM-Pruner)       & 12\% & 0.555 &  8.81 & 0.475 & 14.00 \\
Llama-8B (pruning only)     & 12\% & 0.599 & 10.06 & 0.517 & 11.46 \\
Llama-8B (\koffseq)         & 12\% & 0.598 &  8.79 & 0.540 & 11.36 \\
Llama-8B (\koffjoint)       & 12\% & 0.590 &  8.60 & 0.537 & 10.91 \\
Llama-8B (LLM-Pruner)       & 20\% & 0.471 &  9.98 & 0.396 & 18.72 \\
Llama-8B (pruning only)     & 20\% & 0.400 & 10.88 & 0.532 & 14.03 \\
Llama-8B (\koffseq)         & 20\% & 0.592 &  9.13 & 0.534 & 11.95 \\
Llama-8B (\koffjoint)       & 20\% & 0.602 &  8.12 & 0.537 & 10.90 \\
\midrule
Llama-8B-I (unpruned)       & 0\%  & 0.672 &  7.53 & 0.575 & 10.79 \\
Llama-8B-I (LLM-Pruner)     & 20\% & 0.508 & 10.63 & 0.428 & 20.86 \\
Llama-8B-I (pruning only)   & 20\% & 0.603 &  9.95 & 0.531 & 16.75 \\
Llama-8B-I (\koffseq)       & 20\% & 0.618 &  9.37 & 0.445 & 11.70 \\
Llama-8B-I (\koffjoint)     & 20\% & 0.636 &  9.07 & 0.459 & 12.21 \\
\midrule
Qwen-14B (unpruned)         & 0\%  & 0.779 & 6.26 & 0.741 &  8.33 \\
Qwen-14B (LLM-Pruner)       & 20\% & 0.704 & 8.41 & 0.629 & 12.14 \\
Qwen-14B (pruning only)     & 20\% & 0.734 & 7.58 & 0.712 & 10.56 \\
Qwen-14B (\koffseq)         & 20\% & 0.745 & 6.90 & 0.729 &  8.86 \\
Qwen-14B (\koffjoint)       & 20\% & 0.747 & 6.98 & 0.721 &  8.95 \\
\bottomrule
\end{tabular}}
\caption{Topic- and language-domain results at 12\% and 20\% global sparsity. Pruning only uses the same $L_0$ sparsification and distillation objective but no memory modules. \koffseq learns a mask in a first stage and then trains the memory modules on top of it; \koffjoint learns the mask and the modules together. All systems are trained for one pass over the domain corpus in total. Global MMLU covers the six training languages, each evaluated with its matched module. Qwen-3B is reported only at 12\% sparsity, since at 20\% the memory modules no longer compensate for pruning}
\label{tab:main-and-lang-results}
\end{table*}

\section{Backbone probe protocol}
\begin{figure}[t]
    \centering
    \includegraphics[width=0.5\linewidth]{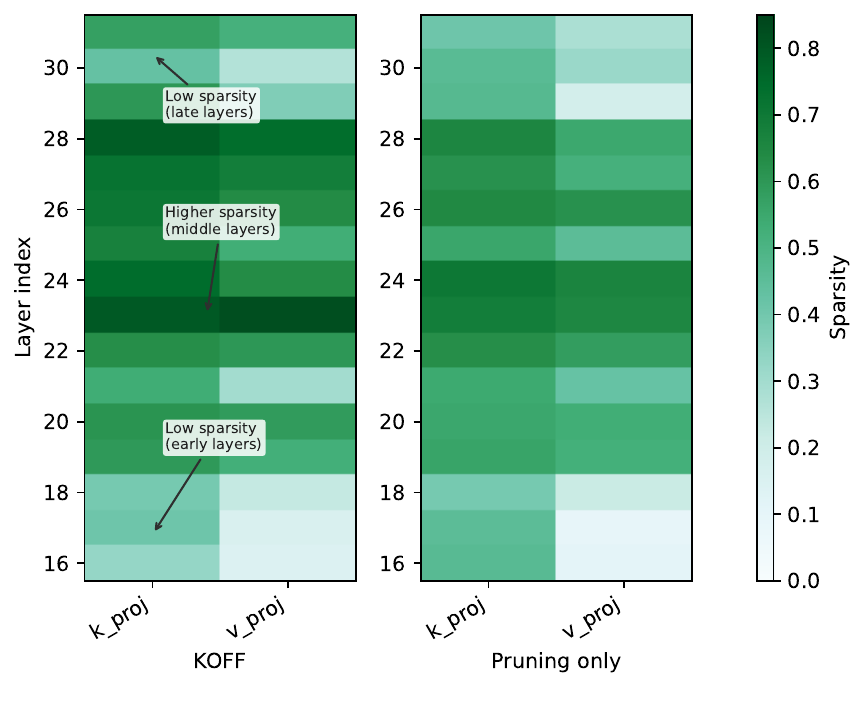}
    \vspace{-0.5cm}
    \caption{Per-layer sparsity for \texttt{k\_proj} and \texttt{v\_proj} layers of Llama-3.1-8B (language domains).}
    \vspace{-0.35cm}
    \label{fig:llama8b_kv_sparsity}
\end{figure}
\label{app:backbone-probes}
Backbone probes are run with adapters disabled and no memory module injection, so all measurements target only the shared pruned backbone. We use the same six-topic Wikipedia-Topics partition as in the main experiments (Mathematics, Science, Technology, History, Culture, Geography), sample up to 5000 texts per topic, and split them 80/20.
   
For representation similarity we report linear CKA~\citep{kornblith2019similarity} between base and pruned test embeddings, along with two neighborhood/cluster diagnostics (kNN-overlap and class-centroid cosine similarity). These diagnostics are intended to check whether pruning preserves global embedding geometry.

\section{Distillation details}
\label{app:distillation}

Table~\ref{tab:distillation-detail} compares three training objectives on Llama-3.2-3B with topic domains: next-token cross-entropy (SFT), KL to the teacher's top-100 logits (our default), and an equal mix of the two ($\alpha{=}0.5$). KL gives higher MMLU and BBL than SFT at similar perplexity. The mixed objective diverges, as it also did in preliminary runs on Qwen-2.5-3B and Llama-3.1-8B; a possible cause is that the two terms conflict once pruning has shifted the student's distribution away from the teacher's.

\section{Memory module placement in the model and pruning targets}
\label{app:placement}

We study two design choices on Llama-3.2-3B with topic domains at a $\sim$12\% global sparsity target: (a) which weight matrices the shared mask $z$ acts on (Section~\ref{sec:where-to-prune}), and (b) which layers the memory modules are attached to (Section~\ref{sec:where-to-attach}).

\begin{table*}[t]
\centering
\footnotesize
\begin{tabular}{lllcccc}
\toprule
Model & Domain & Loss & MMLU$\uparrow$ & BBL$\uparrow$ & PPL$\downarrow$ & Sparsity \\
\midrule
\multirow{3}{*}{Llama-3.2-3B}
  & Topic & SFT                    & 0.482 & 0.446 & 8.77 & 0.121 \\
  & Topic & Blended ($\alpha{=}0.5$) & 0.304 & 0.350 & 86.42 & 0.125 \\
  & Topic & Teacher-only KL           & 0.520 & 0.470 & 8.85 & 0.120 \\
\bottomrule
\end{tabular}%
\caption{Training-objective comparison on Llama-3.2-3B (topic domains, $\sim$12\% global sparsity). Blended training diverges; teacher-only KL gives the best MMLU.}
\label{tab:distillation-detail}
\end{table*}

\subsection{Which Modules to Prune}
\label{sec:where-to-prune}

On Llama-3.2-3B, the MLP projections (\texttt{gate}, \texttt{up}, \texttt{down}) hold roughly three quarters of the non-embedding weights, so gating only the attention projections cannot reach our 12\% global sparsity target: \texttt{k,v} saturates at 2.2\% and \texttt{q,k,v} at 2.8\%. Pruning the attention projections this hard is also destructive: even at 2--3\% global sparsity, both configurations score worse than gating \texttt{k,v} together with the MLP at 12\% (Table~\ref{tab:module-anatomy}).

We leave \texttt{o\_proj} out: because it writes directly into the residual stream, gating its outputs corrupts the shared representation and collapses quality. As a single-run sanity check on Llama-3.2-3B, adding \texttt{o\_proj} to the gated set under our default configuration drops MMLU to near-random, against 0.520 for the matched \texttt{k,v}+MLP setup of Table~\ref{tab:module-anatomy}; we therefore exclude it from all reported configurations.

We likewise leave \texttt{q\_proj} out, as it adds little prunable budget and its benefit is at best marginal and architecture-dependent (on Qwen-2.5-3B the QKV gap nearly closes).
MLP gates converge more slowly than attention gates---about $9{,}644$ versus $1{,}280$ steps to target, so we allow a longer schedule when the MLP is gated.

\begin{table}[t]
\centering
\footnotesize
\setlength{\tabcolsep}{4pt}
\begin{tabular}{lccc}
\toprule
Target Modules & MMLU$\uparrow$ & PPL$\downarrow$ & Sparsity \\
\midrule
\texttt{k,v}+MLP & \textbf{0.520} & \textbf{8.85} & \textbf{0.120} \\
\texttt{k,v} only & 0.488 & 10.26 & 0.022 \\
\texttt{q,k,v} & 0.490 & 10.16 & 0.028 \\
\bottomrule
\end{tabular}
\caption{Module-set comparison on Llama-3.2-3B (topic-domain, rank 128,
$\sim$12\% global-sparsity target). Pruning the MLP is required to reach a meaningful global sparsity: attention-only gating saturates far below target and yields lower MMLU at higher perplexity.}
\label{tab:module-anatomy}
\end{table}

\subsection{Which layers to attach memory modules to}
\label{sec:where-to-attach}

A memory module can in principle be attached at every transformer block, but doing so spreads its parameter budget thinly and lets domain-specific gradients reach early layers whose representations are useful to many domains. To study this, we vary the memory modules placement on Llama-3.2-3B between the upper $N$ blocks of the 28-layer stack and all 28 blocks, holding the rest of the pipeline (\texttt{k,v}+MLP, rank 128, 512 memory tokens, $\sim$12\% global sparsity) fixed. Restricting memory modules to the upper 16 blocks recovers MMLU\,=\,0.520 and PPL\,=\,8.85, against MMLU\,=\,0.470 and PPL\,=\,9.08 for the matched all-layer control: an MMLU gain of +0.050 and a lower perplexity. BBL also improves modestly (0.470 vs.\ 0.438). The interpretation lines up with the capacity-relocation framing: the lower layers carry features that many domains share, so freezing them and concentrating the parameter budget on the specialization-heavy upper half preserves more of the backbone's general reasoning, without sacrificing per-domain performance.

Around this configuration, varying the sparsity weight $\lambda \in \{40, 60, 80\}$ changes MMLU only marginally ($\lambda{=}40$ is slightly ahead of $\lambda{=}80$, as it lands at a marginally lower sparsity), so layer placement remains the dominant design choice.

\end{document}

%% file: math_commands.tex
\usepackage{amsmath,amsfonts,bm}

\def\eqref#1{equation~\ref{#1}}

\def\1{\bm{1}}

\DeclareMathAlphabet{\mathsfit}{\encodingdefault}{\sfdefault}{m}{sl}
\SetMathAlphabet{\mathsfit}{bold}{\encodingdefault}{\sfdefault}{bx}{n}



%% file: custom.bib
@inproceedings{foroutan2022discovering,
  title={Discovering language-neutral sub-networks in multilingual language models},
  author={Foroutan, Negar and Banaei, Mohammadreza and Lebret, R{\'e}mi and Bosselut, Antoine and Aberer, Karl},
  booktitle={Proceedings of the 2022 Conference on Empirical Methods in Natural Language Processing},
  pages={7560--7575},
  year={2022}
}

@inproceedings{tang2024language,
  title={Language-specific neurons: The key to multilingual capabilities in large language models},
  author={Tang, Tianyi and Luo, Wenyang and Huang, Haoyang and Zhang, Dongdong and Wang, Xiaolei and Zhao, Wayne Xin and Wei, Furu and Wen, Ji-Rong},
  booktitle={Proceedings of the 62nd Annual Meeting of the Association for Computational Linguistics (Volume 1: Long Papers)},
  pages={5701--5715},
  year={2024}
}

@misc{ashkboos2024slicegpt,
      title={SliceGPT: Compress Large Language Models by Deleting Rows and Columns},
      author={Saleh Ashkboos and Maximilian L. Croci and Marcelo Gennari do Nascimento and Torsten Hoefler and James Hensman},
      year={2024},
      eprint={2401.15024},
      archivePrefix={arXiv},
      primaryClass={cs.LG},
      url={https://arxiv.org/abs/2401.15024},
}

@misc{hu2025fasp,
      title={FASP: Fast and Accurate Structured Pruning of Large Language Models},
      author={Hanyu Hu and Pengxiang Zhao and Ping Li and Yi Zheng and Zhefeng Wang and Xiaoming Yuan},
      year={2025},
      eprint={2501.09412},
      archivePrefix={arXiv},
      primaryClass={cs.LG},
      url={https://arxiv.org/abs/2501.09412},
}

@misc{ma2023llmpruner,
      title={LLM-Pruner: On the Structural Pruning of Large Language Models},
      author={Xinyin Ma and Gongfan Fang and Xinchao Wang},
      year={2023},
      eprint={2305.11627},
      archivePrefix={arXiv},
      primaryClass={cs.CL},
      url={https://arxiv.org/abs/2305.11627},
}

@misc{louizos2018learning,
      title={Learning Sparse Neural Networks through $L_0$ Regularization},
      author={Christos Louizos and Max Welling and Diederik P. Kingma},
      year={2018},
      eprint={1712.01312},
      archivePrefix={arXiv},
      primaryClass={stat.ML},
      url={https://arxiv.org/abs/1712.01312},
}

@inproceedings{sukhbaatar2015endtoend,
  title={End-To-End Memory Networks},
  author={Sukhbaatar, Sainbayar and Szlam, Arthur and Weston, Jason and Fergus, Rob},
  booktitle={Advances in Neural Information Processing Systems (NeurIPS)},
  volume={28},
  year={2015},
  url={https://arxiv.org/abs/1503.08895},
}

@inproceedings{lewis2020rag,
  title={Retrieval-Augmented Generation for Knowledge-Intensive {NLP} Tasks},
  author={Lewis, Patrick and Perez, Ethan and Piktus, Aleksandra and Petroni, Fabio and Karpukhin, Vladimir and Goyal, Naman and K{\"u}ttler, Heinrich and Lewis, Mike and Yih, Wen-tau and Rockt{\"a}schel, Tim and Riedel, Sebastian and Kiela, Douwe},
  booktitle={Advances in Neural Information Processing Systems (NeurIPS)},
  volume={33},
  year={2020},
  url={https://arxiv.org/abs/2005.11401},
}

@misc{li2021prefix,
      title={Prefix-Tuning: Optimizing Continuous Prompts for Generation},
      author={Xiang Lisa Li and Percy Liang},
      year={2021},
      eprint={2101.00190},
      archivePrefix={arXiv},
      primaryClass={cs.CL},
      url={https://arxiv.org/abs/2101.00190},
}

@inproceedings{lester2021prompt,
  title={The Power of Scale for Parameter-Efficient Prompt Tuning},
  author={Lester, Brian and Al-Rfou, Rami and Constant, Noah},
  booktitle={Proceedings of the 2021 Conference on Empirical Methods in Natural Language Processing (EMNLP)},
  pages={3045--3059},
  year={2021},
  url={https://aclanthology.org/2021.emnlp-main.243/},
}

@misc{pouransari2025hierarchical,
      title={Pretraining with hierarchical memories: separating long-tail and common knowledge},
      author={Hadi Pouransari and David Grangier and C Thomas and Michael Kirchhof and Oncel Tuzel},
      year={2025},
      eprint={2510.02375},
      archivePrefix={arXiv},
      primaryClass={cs.CL},
      url={https://arxiv.org/abs/2510.02375},
}

@inproceedings{hu2022lora,
  title={{LoRA}: Low-Rank Adaptation of Large Language Models},
  author={Hu, Edward J. and Shen, Yelong and Wallis, Phillip and Allen-Zhu, Zeyuan and Li, Yuanzhi and Wang, Shean and Wang, Lu and Chen, Weizhu},
  booktitle={The Tenth International Conference on Learning Representations (ICLR)},
  year={2022},
  url={https://openreview.net/forum?id=nZeVKeeFYf9},
}

@misc{hendrycks2021measuring,
      title={Measuring Massive Multitask Language Understanding},
      author={Dan Hendrycks and Collin Burns and Steven Basart and Andy Zou and Mantas Mazeika and Dawn Song and Jacob Steinhardt},
      year={2021},
      eprint={2009.03300},
      archivePrefix={arXiv},
      primaryClass={cs.CY},
      url={https://arxiv.org/abs/2009.03300},
}

@misc{qwen2025qwen25,
      title={Qwen2 Technical Report},
      author={An Yang and Baosong Yang and Binyuan Hui and Bo Zheng and Bowen Yu and Chang Zhou and Chengpeng Li and Chengyuan Li and Dayiheng Liu and Fei Huang and Guanting Dong and Haoran Wei and Huan Lin and Jialong Tang and Jialin Wang and Jian Yang and Jianhong Tu and Jianwei Zhang and Jianxin Ma and Jianxin Yang and Jin Xu and Jingren Zhou and Jinze Bai and Jinzheng He and Junyang Lin and Kai Dang and Keming Lu and Keqin Chen and Kexin Yang and Mei Li and Mingfeng Xue and Na Ni and Pei Zhang and Peng Wang and Ru Peng and Rui Men and Ruize Gao and Runji Lin and Shijie Wang and Shuai Bai and Sinan Tan and Tianhang Zhu and Tianhao Li and Tianyu Liu and Wenbin Ge and Xiaodong Deng and Xiaohuan Zhou and Xingzhang Ren and Xinyu Zhang and Xipin Wei and Xuancheng Ren and Xuejing Liu and Yang Fan and Yang Yao and Yichang Zhang and Yu Wan and Yunfei Chu and Yuqiong Liu and Zeyu Cui and Zhenru Zhang and Zhifang Guo and Zhihao Fan},
      year={2024},
      eprint={2407.10671},
      archivePrefix={arXiv},
      primaryClass={cs.CL},
      url={https://arxiv.org/abs/2407.10671},
}

@article{raffel2020exploring,
  title={Exploring the Limits of Transfer Learning with a Unified Text-to-Text Transformer},
  author={Raffel, Colin and Shazeer, Noam and Roberts, Adam and Lee, Katherine and Narang, Sharan and Matena, Michael and Zhou, Yanqi and Li, Wei and Liu, Peter J.},
  journal={Journal of Machine Learning Research},
  volume={21},
  number={140},
  pages={1--67},
  year={2020},
  url={http://jmlr.org/papers/v21/20-074.html},
}

@misc{cobbe2021gsm8k,
      title={Training Verifiers to Solve Math Word Problems},
      author={Karl Cobbe and Vineet Kosaraju and Mohammad Bavarian and Mark Chen and Heewoo Jun and Lukasz Kaiser and Matthias Plappert and Jerry Tworek and Jacob Hilton and Reiichiro Nakano and Christopher Hesse and John Schulman},
      year={2021},
      eprint={2110.14168},
      archivePrefix={arXiv},
      primaryClass={cs.LG},
      url={https://arxiv.org/abs/2110.14168},
}

@misc{clark2018think,
      title={Think you have Solved Question Answering? Try ARC, the AI2 Reasoning Challenge},
      author={Peter Clark and Isaac Cowhey and Oren Etzioni and Tushar Khot and Ashish Sabharwal and Carissa Schoenick and Oyvind Tafjord},
      year={2018},
      eprint={1803.05457},
      archivePrefix={arXiv},
      primaryClass={cs.AI},
      url={https://arxiv.org/abs/1803.05457},
}

@misc{tarekziade_wikipedia_topics,
  title        = {Wikipedia-Topics Dataset},
  author       = {Ziad{\'e}, Tarek},
  year         = {2024},
  howpublished = {Hugging Face Datasets},
  url          = {https://huggingface.co/datasets/tarekziade/wikipedia-topics},
  note         = {Wikipedia articles partitioned into 40 root categories},
}

@misc{wikimedia,
  title        = {Wikisource: The Free Library},
  author       = {{Wikimedia Foundation}},
  howpublished = {\url{https://wikisource.org}},
  note         = {Multilingual collection of source texts},
}

@inproceedings{zhang-etal-2022-moefication,
    title = "{M}o{E}fication: Transformer Feed-forward Layers are Mixtures of Experts",
    author = "Zhang, Zhengyan  and
      Lin, Yankai  and
      Liu, Zhiyuan  and
      Li, Peng  and
      Sun, Maosong  and
      Zhou, Jie",
    editor = "Muresan, Smaranda  and
      Nakov, Preslav  and
      Villavicencio, Aline",
    booktitle = "Findings of the Association for Computational Linguistics: ACL 2022",
    month = may,
    year = "2022",
    address = "Dublin, Ireland",
    publisher = "Association for Computational Linguistics",
    url = "https://aclanthology.org/2022.findings-acl.71/",
    doi = "10.18653/v1/2022.findings-acl.71",
    pages = "877--890"
}

@inproceedings{komatsuzaki2023sparse,
  title = {Sparse Upcycling: Training Mixture-of-Experts from Dense Checkpoints},
  author = {Komatsuzaki, Aran and
    Puigcerver, Joan and
    Lee-Thorp, James and
    Riquelme Ruiz, Carlos and
    Mustafa, Basil and
    Ainslie, Joshua and
    Tay, Yi and
    Dehghani, Mostafa and
    Houlsby, Neil},
  booktitle = {International Conference on Learning Representations},
  year = {2023},
  url = {https://openreview.net/forum?id=T5nUQDrM4u}
}

@misc{li2022branchtrainmerge,
  title = {Branch-Train-Merge: Embarrassingly Parallel Training of Expert Language Models},
  author = {Li, Margaret and
    Gururangan, Suchin and
    Dettmers, Tim and
    Lewis, Mike and
    Althoff, Tim and
    Smith, Noah A. and
    Zettlemoyer, Luke},
  year = {2022},
  eprint = {2208.03306},
  archivePrefix = {arXiv},
  primaryClass = {cs.CL},
  url = {https://arxiv.org/abs/2208.03306}
}

@inproceedings{sukhbaatar2024branchtrainmix,
  title = {Branch-Train-MiX: Mixing Expert LLMs into a Mixture-of-Experts LLM},
  author = {Sukhbaatar, Sainbayar and
    Golovneva, Olga and
    Sharma, Vasu and
    Xu, Hu and
    Lin, Xi Victoria and
    Rozi{\`e}re, Baptiste and
    Kahn, Jacob and
    Li, Shang-Wen and
    Yih, Wen-tau and
    Weston, Jason E. and
    Li, Xian},
  booktitle = {First Conference on Language Modeling},
  year = {2024},
  url = {https://openreview.net/forum?id=nqLAuMOF6n}
}

@article{eyuboglu2025cartridges,
  title={Cartridges: Lightweight and general-purpose long context representations via self-study},
  author={Eyuboglu, Sabri and Ehrlich, Ryan and Arora, Simran and Guha, Neel and Zinsley, Dylan and Liu, Emily and Tennien, Will and Rudra, Atri and Zou, James and Mirhoseini, Azalia and others},
  journal={arXiv preprint arXiv:2506.06266},
  year={2025}
}

@inproceedings{kornblith2019similarity,
  title={Similarity of neural network representations revisited},
  author={Kornblith, Simon and Norouzi, Mohammad and Lee, Honglak and Hinton, Geoffrey},
  booktitle={International conference on machine learning},
  pages={3519--3529},
  year={2019},
  organization={PMLR}
}


%% file: iclr2027_conference.bib
@article{lin2025continual,
  title={Continual learning via sparse memory finetuning},
  author={Lin, Jessy and Zettlemoyer, Luke and Ghosh, Gargi and Yih, Wen-Tau and Markosyan, Aram and Berges, Vincent-Pierre and O{\u{g}}uz, Barlas},
  journal={arXiv preprint arXiv:2510.15103},
  year={2025}
}

@article{pfeiffermodular,
  title={Modular Deep Learning},
  author={Pfeiffer, Jonas and Ruder, Sebastian and Vuli{\'c}, Ivan and Ponti, Edoardo},
  journal={Transactions on Machine Learning Research},
  year = 2023
}

@inproceedings{pfeiffer2020mad,
  title={Mad-x: An adapter-based framework for multi-task cross-lingual transfer},
  author={Pfeiffer, Jonas and Vuli{\'c}, Ivan and Gurevych, Iryna and Ruder, Sebastian},
  booktitle={Proceedings of the 2020 conference on empirical methods in natural language processing (EMNLP)},
  pages={7654--7673},
  year={2020}
}

@inproceedings{choenni2024examining,
  title={Examining modularity in multilingual lms via language-specialized subnetworks},
  author={Choenni, Rochelle and Shutova, Ekaterina and Garrette, Dan},
  booktitle={Findings of the Association for Computational Linguistics: NAACL 2024},
  pages={287--301},
  year={2024}
}

@article{zhao2025pre,
  title={Pre-training Limited Memory Language Models with Internal and External Knowledge},
  author={Zhao, Linxi and Zalouk, Sofian and Belardi, Christian K and Lovelace, Justin and Zhou, Jin Peng and Noonan, Ryan Thomas and Go, Dongyoung and Weinberger, Kilian Q and Artzi, Yoav and Sun, Jennifer J},
  journal={ICLR},
  year={2026}
}

@article{
srivastava2023beyond,
title={Beyond the Imitation Game: Quantifying and extrapolating the capabilities of language models},
author={Aarohi Srivastava and Abhinav Rastogi and Others},
journal={Transactions on Machine Learning Research},
issn={2835-8856},
year={2023},
url={https://openreview.net/forum?id=uyTL5Bvosj},
note={Featured Certification}
}

@inproceedings{
loshchilov2018decoupled,
title={Decoupled Weight Decay Regularization},
author={Ilya Loshchilov and Frank Hutter},
booktitle={International Conference on Learning Representations},
year={2019},
url={https://openreview.net/forum?id=Bkg6RiCqY7},
}

@article{liu2019roberta,
    title = {RoBERTa: A Robustly Optimized BERT Pretraining Approach},
    author = {Yinhan Liu and Myle Ott and Naman Goyal and Jingfei Du and
              Mandar Joshi and Danqi Chen and Omer Levy and Mike Lewis and
              Luke Zettlemoyer and Veselin Stoyanov},
    journal={arXiv preprint arXiv:1907.11692},
    year = {2019},
}

@inproceedings{singh2025globalmmluunderstandingaddressing,
  title={Global mmlu: Understanding and addressing cultural and linguistic biases in multilingual evaluation},
  author={Singh, Shivalika and Romanou, Angelika and Fourrier, Cl{\'e}mentine and Adelani, David Ifeoluwa and Ngui, Jian Gang and Vila-Suero, Daniel and Limkonchotiwat, Peerat and Marchisio, Kelly and Leong, Wei Qi and Susanto, Yosephine and others},
  booktitle={Proceedings of the 63rd Annual Meeting of the Association for Computational Linguistics (Volume 1: Long Papers)},
  pages={18761--18799},
  year={2025}
}

@misc{grattafiori2024llama3herdmodels,
      title={The Llama 3 Herd of Models}, 
      author={Aaron Grattafiori and Abhimanyu Dubey and Abhinav Jauhri and Abhinav Pandey and Abhishek Kadian and Ahmad Al-Dahle and Aiesha Letman and Akhil Mathur and Others},
      year={2024},
      eprint={2407.21783},
      archivePrefix={arXiv},
      primaryClass={cs.AI},
      url={https://arxiv.org/abs/2407.21783}, 
}
